\documentclass{bmvc2k}
\usepackage{subfigure}
\usepackage{graphicx}
\usepackage{tabulary}
\usepackage{placeins} 

\title{DeepLogo: Hitting Logo Recognition with the Deep Neural Network Hammer}

\addauthor{~~~~Forrest N. Iandola}{forresti@eecs.berkeley.edu}{1}
\addauthor{~~~~Anting Shen}{antingshen@berkeley.edu}{1}
\addauthor{~~~~Peter Gao}{pgao@berkeley.edu}{1}
\addauthor{~~~~Kurt Keutzer}{keutzer@eecs.berkeley.edu}{1}

\addinstitution{
  ASPIRE Laboratory \\
 Unviersity of California, Berkeley, CA, USA
}

\runninghead{Iandola, Shen, Gao, Keutzer}{DeepLogo}

\begin{document}

\maketitle
\vspace{-0.2in}
\begin{abstract}
Recently, there has been a flurry of industrial activity around logo recognition, such as Ditto's service for marketers to track their brands in user-generated images, and LogoGrab's mobile app platform for logo recognition.
However, relatively little academic or open-source logo recognition progress has been made in the last four years.
Meanwhile, deep convolutional neural networks (DCNNs) have revolutionized a broad range of object recognition applications.
In this work, we apply DCNNs to logo recognition.
We propose several DCNN architectures, with which we surpass published state-of-art accuracy on a popular logo recognition dataset.
\end{abstract}
\vspace{-0.1in}

\section{Introduction}
\vspace{-0.1in}
\label{sec:intro}

Logo recognition is the key problem in contextual ad placement, validation of product placement, and online brand management.
{\em Contextual} ad placement is about placing relevant ads on webpages, images, and videos.
An obvious problem in contextual advertising is "what brands appear in images and video content?" 
Logo recognition is the core solution to this.

As absurd as it may sound, ad networks often don't know what's in the ad JPEGs that they are placing in front of web content\footnote{\bf Managers at three different digital advertising platforms told us about this problem.}.
For example, a brand like Toyota can import an ad into Google AdWords without mentioning that "this is a Toyota ad, and we may want to place this on car-related content."
We have observed that most ads contain their brand's logo, so logo recognition is the obvious solution to this.
We provide a deeper explanation of these logo recognition applications in Section~\ref{sec:applications}.

In this paper, we apply deep convolutional neural networks (DCNNs) to logo recognition.
We look at three problem statements in logo recognition: classification, detection without localization, and detection with localization.
We achieve state-of-the-art accuracy by customizing DCNNs for logo recognition.

The rest of the paper is organized as follows.
In Section~\ref{sec:related}, we describe related work in logo recognition methods, logo recognition applications, and object recognition.
Section~\ref{sec:data} provides an overview of publicly-available datasets of labeled logo images.
In Section~\ref{sec:architectures}, we propose neural network architectures for logo recognition.
In Section~\ref{sec:eval}, we evaluate our classifiers in logo classification and detection.
We conclude in Section~\ref{sec:conclusion}.

\section{Related Work}
\vspace{-0.1in}
\label{sec:related}

\subsection{Logo Recognition Methods}

\vspace{-0.1in}

In the last few years (2007-2014), Scale Invariant Feature Transform (SIFT) features have been at the core of most logo recognition methods, replacing methods such as Fisher classifiers \cite{Guangyu2007}.
Many previous works employ a bag-of-words framework to work with SIFT features,
where local features are quantized into a vocabulary to allow for efficient matching of descriptors.
Boia et al. \cite{Boia2014} applied Complete Rank Transform on top of bag-of-words SIFT features to gain additional invariance to illumination and monotonic changes in pixel intensities in their logo classifier, and achieve state of the art classification accuracies that we use as our baseline. 
For detection, Romberg et al. \cite{Romberg2011} combined a cascaded index with bag-of-words SIFT features to achieve high precision at the expense of some recall while maintaining computational efficiency and scalability. 

Neural networks have played a part in logo detection as well.
Psyllos et al. \cite{Psyllos2011} incorporated neural networks by applying a probabilistic neural network on top of SIFT for vehicle model recognition. 
Francesconi\cite{Francesconi1998} used recursive neural networks to classify black-and-white logos. 
Duffner \& Garcia \cite{duffner2006} fed pixel values directly into a  convolutional neural network with two convolution layers to detect watermarks on television.

\subsection{Logo Recognition Applications}
\vspace{-0.1in}
\label{sec:applications}
In this section, we summarize real-world applications where logo recognition plays a crucial role.

\begin{table}[htb]
	\footnotesize
	\caption{Applications of logo classification and detection.}
	\label{T:applications}
	\centering
	\begin{tabular}{|p{2.5cm}|p{4cm}|p{4cm}|}
		\hline
		& Problem Definition                                   & Example Applications  \\ \hline
		Logo Classification                      & Which logo is in this image?                    & Checking for mistakes in used-car ads~\cite{Psyllos2011} \\ \hline
		Logo Detection w/o Localization  & If there is a logo, what type of logo is it? & Marketing, brand tracking on Instagram and Pinterest~\cite{Ditto}; Document classification~\cite{Guangyu2007}.  \\ \hline
		Logo Detection with Localization & Locate and classify logos in the image.   & Augmented reality~\cite{LogoGrab}  \\ \hline
		Logo Localization                        & Locate the logos, but don't classify them. & Removing watermarks from television content~\cite{duffner2006}~~\cite{Yan2005} \\ \hline
	\end{tabular}
\end{table}

{\bf Brand tracking on Instagram.} Since the rise of social media in the mid-2000s, companies like HootSuite and Sprout Social have provided dashboards for marketers to track their brands on Facebook and Twitter.
Most of these social media analytics dashboards simply aggregate text data from tweets and Facebook posts.
However, Instagram and Pinterest are social networks in which communications are based almost entirely on photos.
Logo recognition is key to brand analytics on Instagram and Pinterest.

{\bf Used car ad verification.} Psyllos et al. used logo recognition (e.g. Ford, Volkswagen, etc) as one of several signals for classifying cars by make and model~\cite{Psyllos2011}. 
Their logo recognition system consists of SIFT features classified with a Probabilistic Neural Network. 
This would be useful for identifying inaccurate listings on eBay or Autotrader used car classified ads -- e.g., flagging an ad labeled as a Honda CR-V that contains photos of a Volkswagen Passat.

{\bf Watermark removal.} Yan et al.~\cite{Yan2005} and Duffner et al.~\cite{duffner2006} used logo recognition as part of a system for removing logos and watermarks from television content. 
Note that this application does not necessarily need to identify the {\em type} of logo (e.g. Apple or Adidas); it simply needs to localize logos and remove them. 

We show how each of these applications uses logo recognition in Table~\ref{T:applications}. 
Note that there are a few variants of the logo recognition problem, such as {\em classification} and {\em detection.}

\subsection{Deep Neural Networks for Object Recognition}
\vspace{-0.1in}

In our view, logo recognition is an instantiation of the broader problem of {\em object recognition}. 
Recently, Deep Convolutional Neural Networks (DCNNs) have unleashed a torrent of progress on object recognition.
In the ImageNet object classification challenge, DCNNs have posted accuracy improvements of several percentage points per year.
Using DCNNs, Donohue et al. outperformed state-of-the-art accuracy on scene classification and fine-grained bird classification~\cite{decaf}.
DCNNs enabled Razavian et al. to outperform state-of-the-art accuracy human attribute detection and visual instance retrieval~\cite{Razavian2014}.

\vspace{-0.1in}
\section{Data}
\vspace{-0.1in}
\label{sec:data}

\begin{table}[htb]
	
	\caption{Overview of the {\bf FlickrLogos-32} dataset. Foreground images contain logos; background images do not.}
	\label{T:data_overview}
	\centering
	\begin{tabular}{|c|c|c|}
		\hline
		& trainval & test \\ \hline
		foreground & 1280 & 960 \\ \hline
		background & 3000 & 3000 \\ \hline
	\end{tabular}
\end{table}

There are several datasets of images labeled with logos, including BelgaLogos~\cite{Joly09}, FlickrLogos-27~\cite{Kalantidis2011}, and FlickrLogos-32~\cite{Romberg2011}.
We evaluate our logo detection methods on the FlickrLogos-32 dataset, because it has more labeled logo images than any other dataset that we are aware of.
We give an overview of FlickrLogos-32 in Table~\ref{T:data_overview}.
We show a few example images from the FlickrLogos-32 dataset in Figure~\ref{fig:flickrlogos_examples}.

\begin{figure}[htb]
\centering
      \begin{tabular}{ccc}
        \bmvaHangBox{\fbox{\includegraphics[height=25mm]{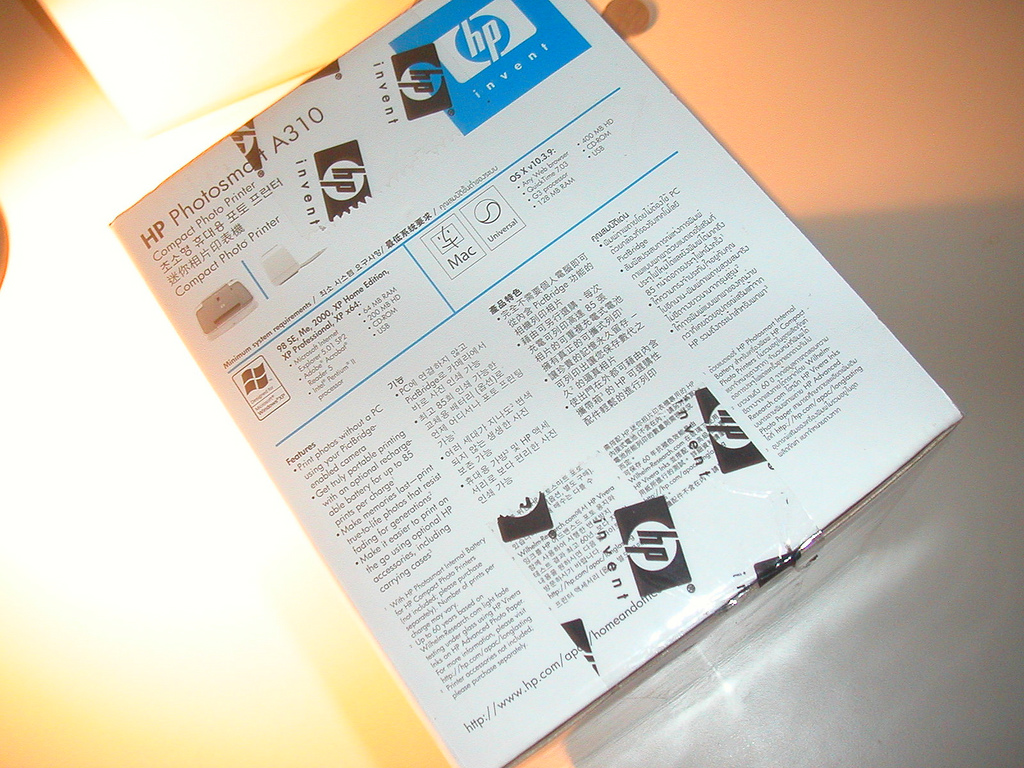}}}&
        \bmvaHangBox{\fbox{\includegraphics[height=25mm]{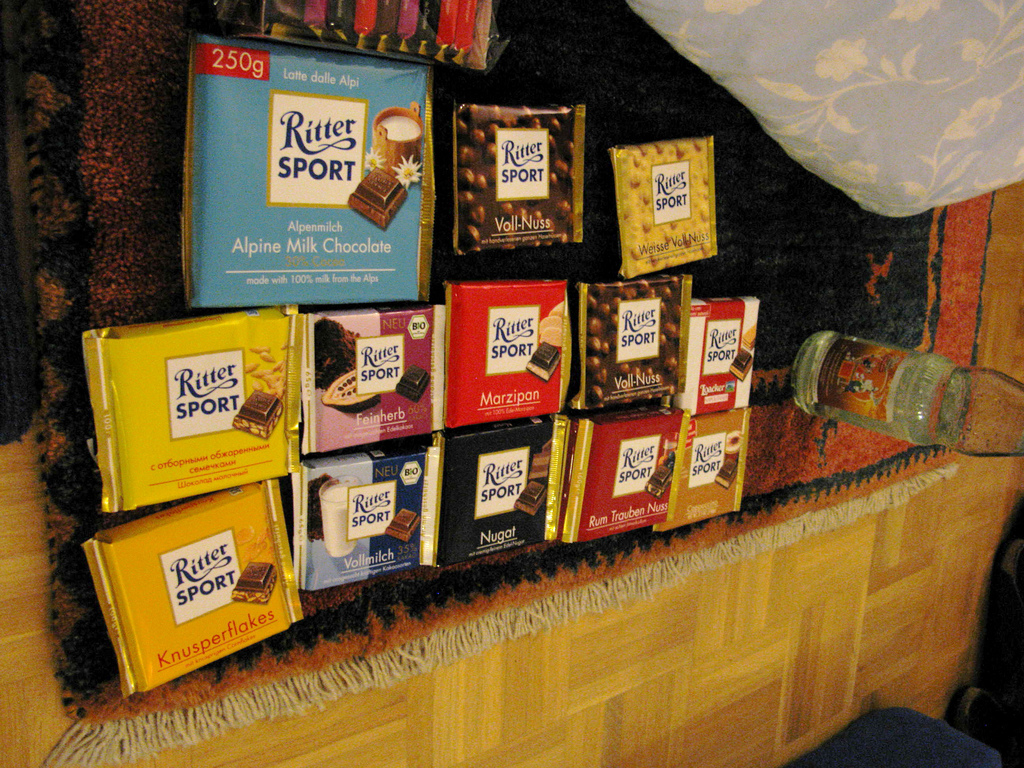}}}&
        \bmvaHangBox{\fbox{\includegraphics[height=25mm]{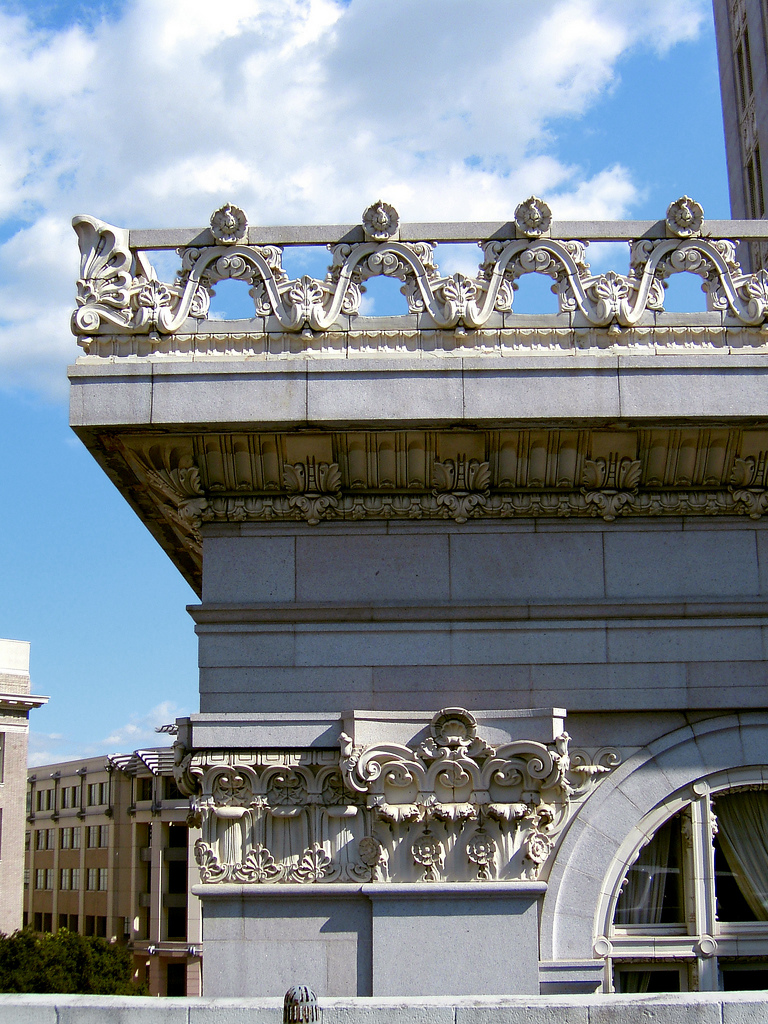}}} \\
        (a) & (b) & (c) \\
        HP & Ritter Sport & Background image (no logos) \\
      \end{tabular}
    \caption{{\bf Diversity of the FlickrLogos-32 dataset.} The logos vary widely in size, down to just a few pixels (a). Some images like (b) have multiple images of the same type. However, an image wouldn't contain both Ritter Sport and HP. "Background images" like (c) don't contain logos.}
    \label{fig:flickrlogos_examples}
\end{figure}

The FlickrLogos-32 dataset also provides bounding box annotations for all logos, which can optionally be used during training.
In Sections~\ref{sec:classification} and~\ref{sec:detection_without_localization}, we ignore the bounding boxes.
We use the bounding boxes in Section~\ref{sec:detection_with_localization}.

\FloatBarrier
\section{Deep Convolutional Neural Network Architectures}
\vspace{-0.1in}
\label{sec:architectures}

\begin{figure}[htb]
\centering
    \fbox{\includegraphics[height=30mm]{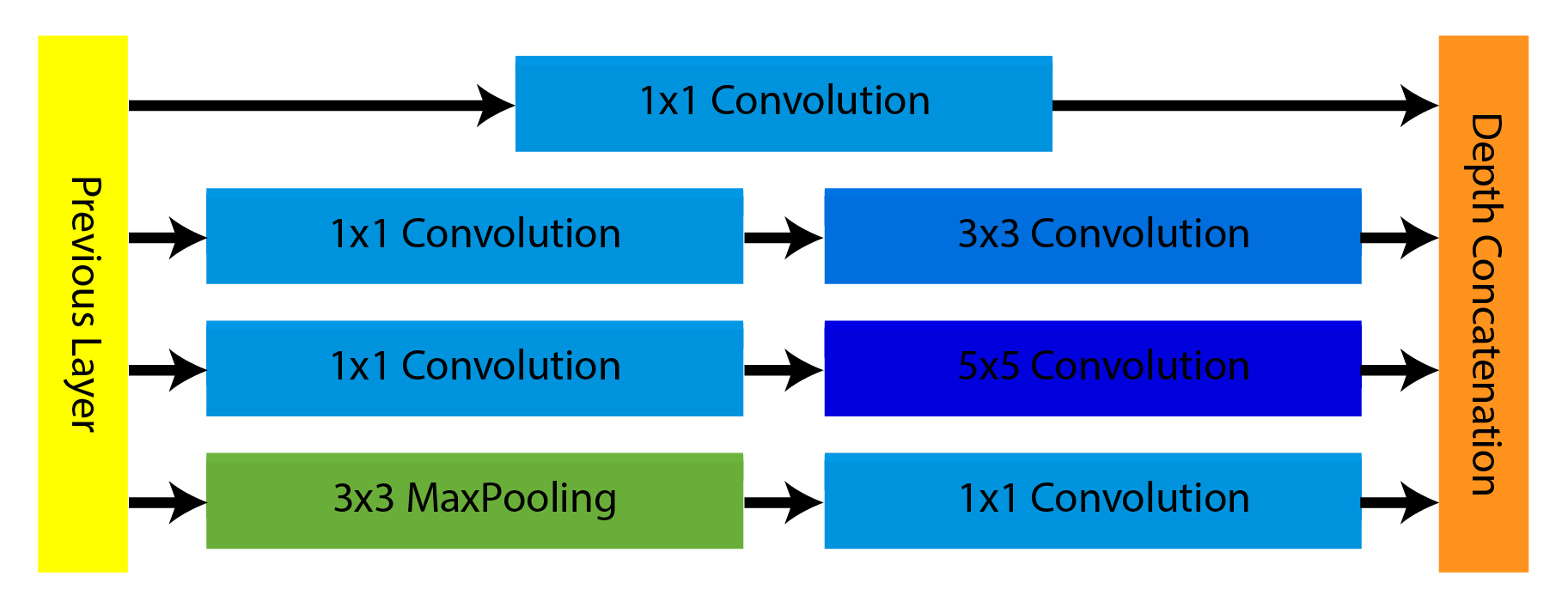}}
    \caption{An ``inception meta-layer," as defined in GoogLeNet.}
    \label{fig:inception_layer}
\end{figure}

\begin{figure}[htb]
\centering
    \fbox{\includegraphics[height=50mm]{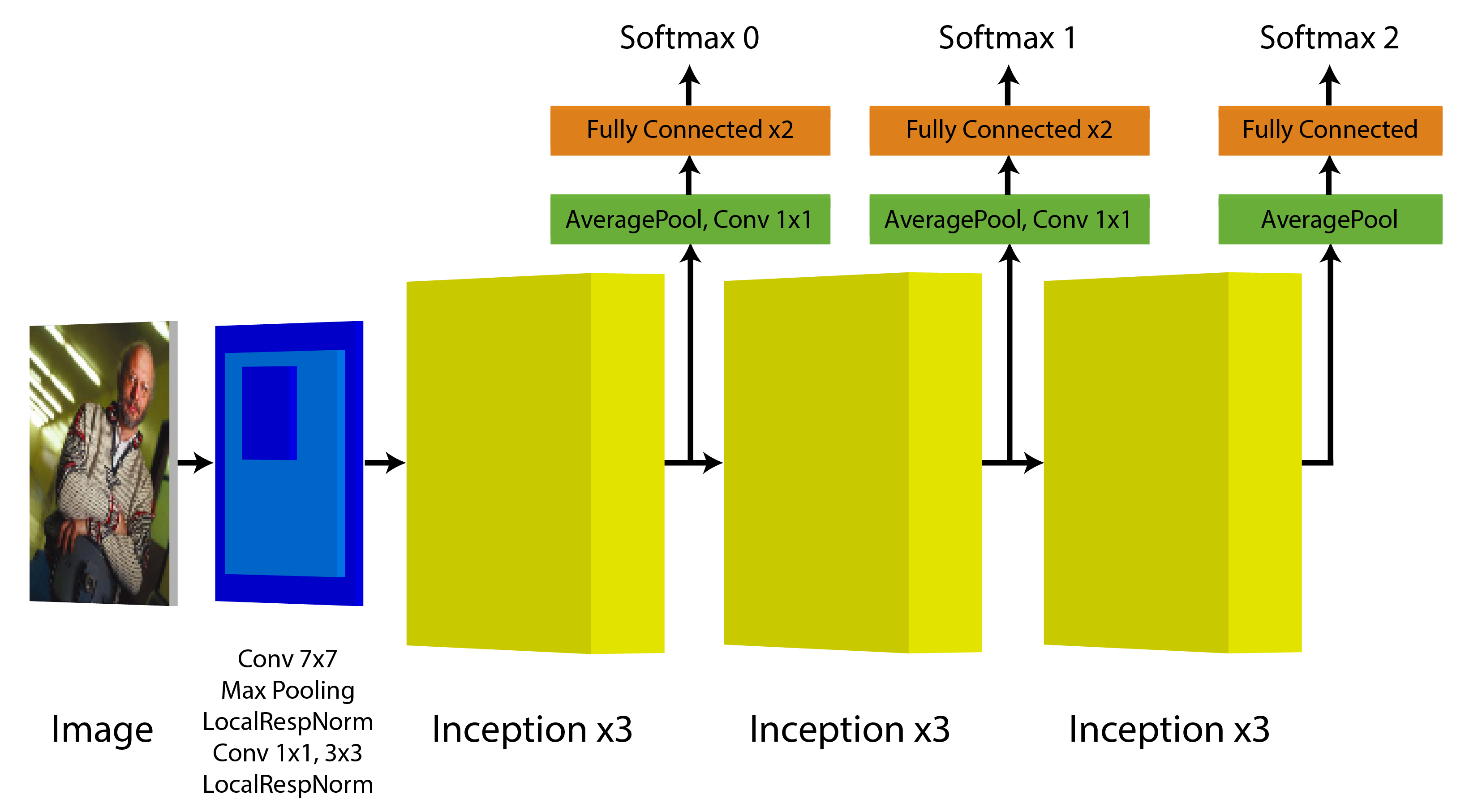}}
    \caption{{\bf GoogLeNet} consists almost entirely of `inception meta-layers."}
    \label{fig:googlenet}
\end{figure}

We now review DCNN architectures from the literature, which we will use as inspiration to design specialized architectures for logo recognition.
In 2012, Krizhevsky et al. designed AlexNet~\cite{alexnet}, a DCNN architecture containing 5 convolutional layers and 3 fully-connected layers. 
AlexNet won the ImageNet-2012 image classification challenge.
By 2014, state-of-the-art ImageNet DCNNs had evolved into extremely deep networks with up to 19 layers, such as VGG-19~\cite{VGG-19}.
In addition to increasing depth, GoogLeNet (Figure~\ref{fig:googlenet}) is comprised of "inception" meta-layers (Figure~\ref{fig:inception_layer}), which contain multiple convolution filter resolutions. 
Inception layers are well equipped to classify images where objects may be large or small with respect to the overall image size.
GoogLeNet and VGG-19 took the top two places in the ImageNet-2014 competition.
We give a broader context of the speed/accuracy tradeoffs among these DCNN architectures in Table~\ref{T:imagenet-accuracy}.

\begin{table}[htb]
\scriptsize
\caption{{\bf ImageNet accuracy} for popular deep convolutional neural network architectures. Single-crop accuracy numbers are reported by the respective authors. We obtained these per-frame speed and energy numbers on a NVIDIA K40 GPU using Caffe's native convolution implementation~\cite{jia2014caffe}.}
\label{T:imagenet-accuracy}
\centering
\begin{tabular}{|c|c|c|c|c|c|c|}
\hline
DCNN Architecture &  Top-5 ImageNet-1k accuracy &  Speed (Caffe) & Energy (Caffe) \\ \hline
AlexNet~\cite{alexnet} & 81.8\% & 417 fps & 0.48 kJ/frame \\ \hline
VGG-19~\cite{VGG-19}  & 91.0\% & 35.1 fps & 8.37 kJ/frame \\ \hline   
GoogLeNet~\cite{googlenet} & \bf 89.9\% & 121 fps & 1.64 kJ/frame \\ \hline 
\end{tabular}
\end{table}

\subsection{Our Proposed DCNN architectures}
\vspace{-0.1in}
GoogLeNet is a good start, but we believe that there are even better DCNN architectures for logo classification.
Within the vast DCNN architectural space, we are on the hunt for architectures that are even better than GoogLeNet at handling a broad variety of logo resolutions (see Figure~\ref{fig:logo-scale}).
Here are three examples of architectures that we have designed for this goal:

{\bf GoogLeNet-GP}, or GoogLeNet with Global Pooling, performs global average-pooling prior to fully connected layers. 
We implement this with the Caffe global pooling layer, which takes {\em any} input layer and pools it down to 1$x$1$x$channels.
This enables the network to process any input image size at training or test time.
We exploit this property of GoogLeNet-GP by feeding in various image sizes at training time.

{\bf GoogLeNet-FullClassify}. 
When making nets extremely deep, end-to-end backpropagation can become ineffective, as discussed in~\cite{VGG-19}.
One way to make a very deep net more trainable with backpropagation is to have multiple softmax classifiers hanging off the side of the net.
For example, GoogLeNet has 3 softmax classifiers, situated after Inception3, Inception6, and Inception9.
Taking this idea to an extreme, we propose GoogLeNet-FullClassify, which has a classifier after every inception layer.
At test time, we only use the final softmax layer's classification outputs.

{\bf Full-Inception}. 
A "Full-Inception" net is like GoogLeNet, but the first layer is also an inception layer.
Most DCNNs, including GoogLeNet, struggle to classify low-resolution logos in big images.
Full-Inception aims to rectify this by having a range of DCNN filter sizes at the first layer, which operates on the input pixels.

\section{Evaluation}
\vspace{-0.1in}
\label{sec:eval}

We now evaluate deep convolutional neural networks on the FlickrLogos-32 logo recognition dataset -- we train on {\tt trainval} and test on {\tt test}. 
All of our experiments were conducted with the Caffe~\cite{jia2014caffe} deep learning framework. 
We consider three variants of the logo recognition problem: classification, detection without localization, and detection with localization.

\subsection{Logo Classification}
\vspace{-0.1in}
\label{sec:classification}

\begin{table}[htb]
\footnotesize
\caption{Results for {\bf logo classification} on FlickrLogos-32 with no background samples. Pretraining: ImageNet (ILSVRC-2012-train); Fine-tuning: FlickrLogos-32-trainval; Testing: FlickrLogos-32-test. In ``GoogLeNet-GP," we average pool down to 1$x$1$x$channels (aka ``globalpool") prior to the fc layers.}
\label{T:logo-classification}
\centering
\begin{tabular}{|p{2.5cm}|c|c|c|c|}
\hline
Method & Dropout & Pretrain & Finetune & Accuracy\\ \hline
Random Chance & - & - & FlickrLogos & $\frac{1}{32}$ = 3.12\% \\ \hline
baseline: Complete Rank Transform~\cite{Boia2014} & - & - & FlickrLogos & 88.97\% \\ \hline
VGG-19 (ours) & 0.5 & ImageNet-2012 & FlickrLogos & 3.12\% \\ \hline
AlexNet (ours) & 0.5 & ImageNet-2012 & FlickrLogos & 70.1\% \\ \hline
Full-Inception (ours) & 0.7 & ImageNet-2012 & FlickrLogos & 77.1\% \\ \hline 
GoogLeNet (ours) & 0.7 & ImageNet-2012 & FlickrLogos & 87.6\% \\ \hline
GoogLeNet (ours) & 0.8 & ImageNet-2012 & FlickrLogos & 88.7\% \\ \hline
GoogLeNet-GP (ours) & 0.8 & ImageNet-2012 & FlickrLogos & 89.1\% \\ \hline
GoogLeNet-GP (ours) & 0.9 & ImageNet-2012 & FlickrLogos & {\bf 89.6\%} \\ \hline
FullClassify (ours) & 0.9 & ImageNet-2012 & FlickrLogos & 89.2\% \\ \hline
\end{tabular}
\end{table}
We now turn to the problem of classifying logos in images.
As we describe in Table~\ref{T:applications}, we define this problem as "Which logo is in this image?"
In this section, we assume that each image in the test set contains a logo.

\begin{figure}[htb]
  \begin{center}
    \includegraphics[height=50mm]{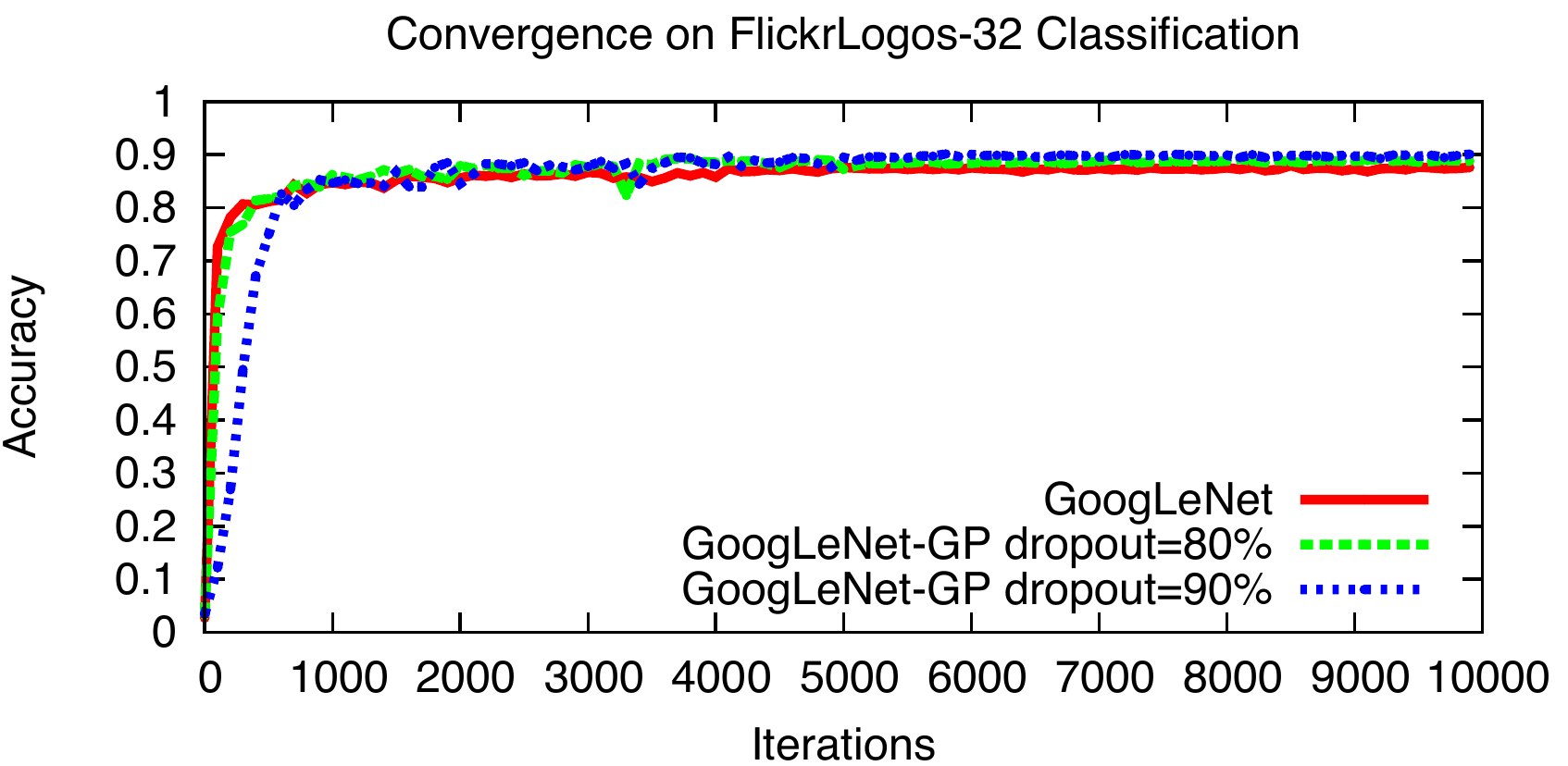} 

  \end{center}
    \caption{{\bf Convergence:} fine-tuning our variants of GoogLeNet for FlickrLogos-32 classification with a batch size of 32. As we increase dropout, training takes slightly longer to converge. The training has mostly converged by 1000 iterations. On an NVIDIA K20 GPU, 1000 iterations of GoogLeNet (or GoogLeNet-GP) training takes 15 minutes and 32 seconds.}
    
   \label{fig:convergence}
\end{figure}

\begin{figure}[htb]
	\begin{center}
        \includegraphics[height=50mm]{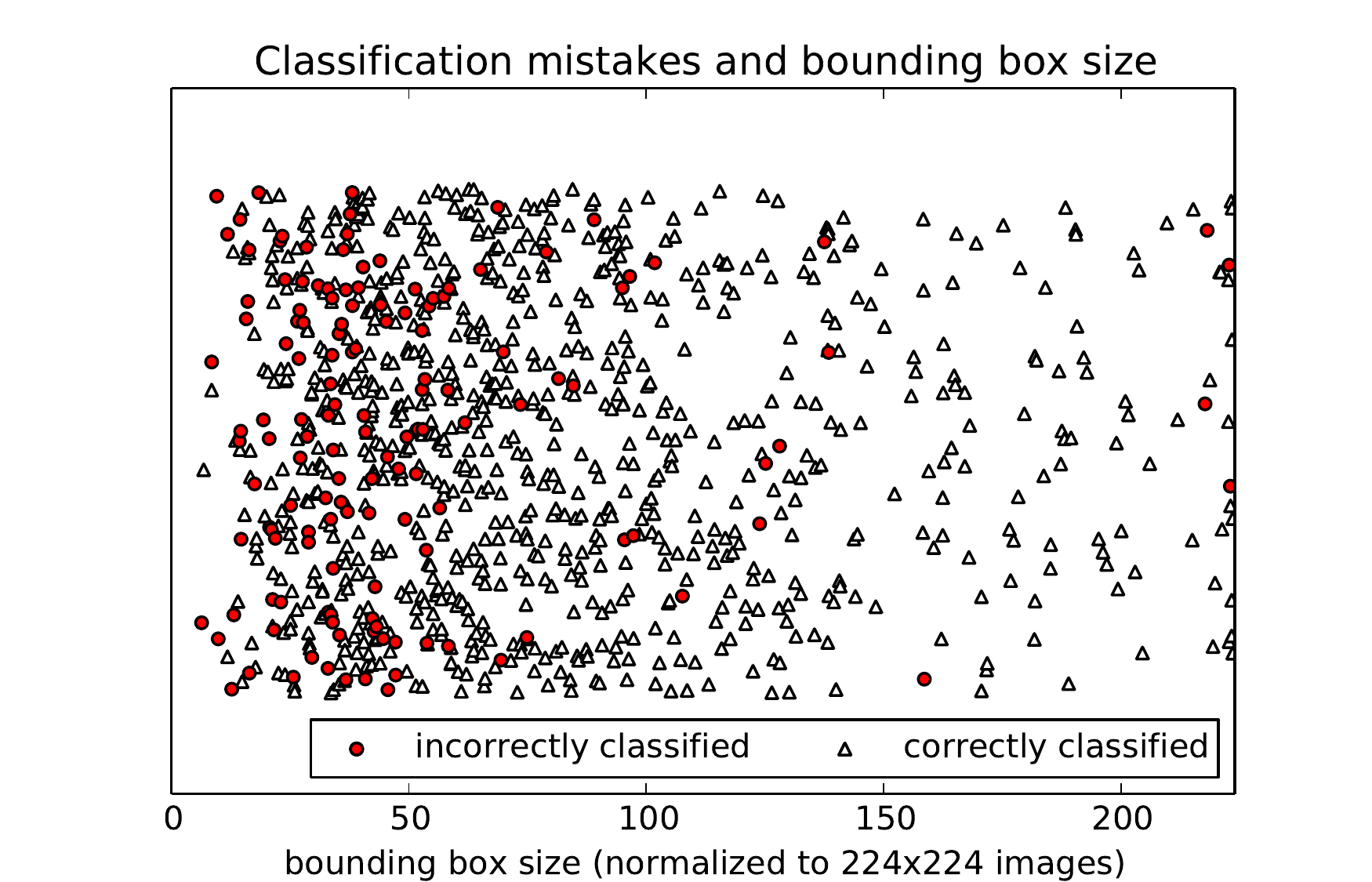} 
        \vspace{-0.1in}
	\end{center}
	\caption{Correlation between bounding box size and logo classification outcome for {\bf GoogLeNet-GP, dropout=0.9}. Failures are more common for smaller logos.}
	
	\label{fig:mistakes_bbox_size}
\end{figure}

\begin{figure}[htb]
	\begin{center}
		\begin{tabular}{cccc}
			\bmvaHangBox{\fbox{\includegraphics[height=40mm]{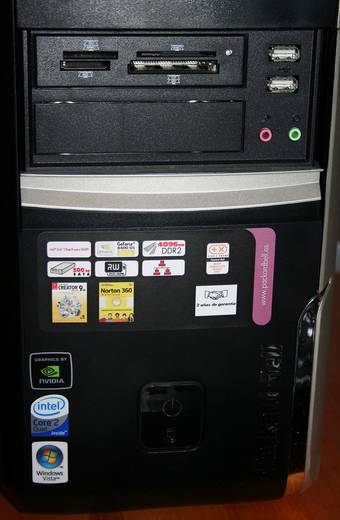}}}&
			\bmvaHangBox{\fbox{\includegraphics[height=40mm]{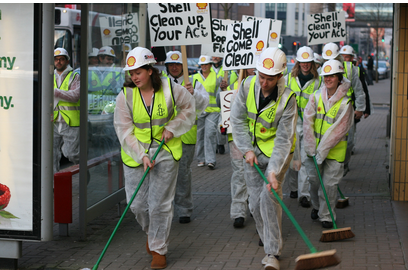}}}\\
			(a) & (b)\\
			NVIDIA & Shell\\
		\end{tabular}
		\caption{{\bf Failure case: low-resolution logos.} We misclassified photos like (a) and (b) because the logos are small in relation to the image resolution. }
		\label{fig:logo-scale}
	\end{center}
\end{figure}

\begin{figure}[htb]
  \begin{center}
  \begin{tabular}{cccc}
    \bmvaHangBox{\fbox{\includegraphics[height=20mm]{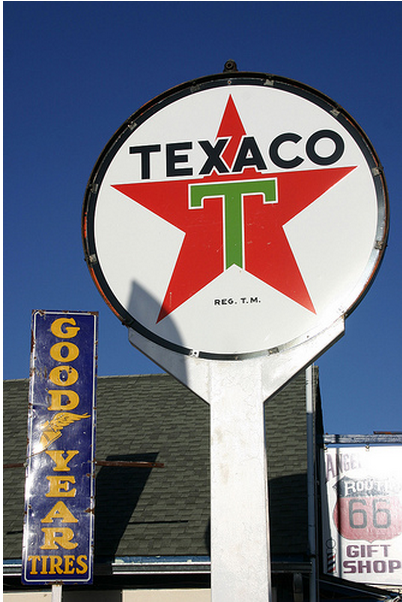}}}&
    \bmvaHangBox{\fbox{\includegraphics[height=20mm]{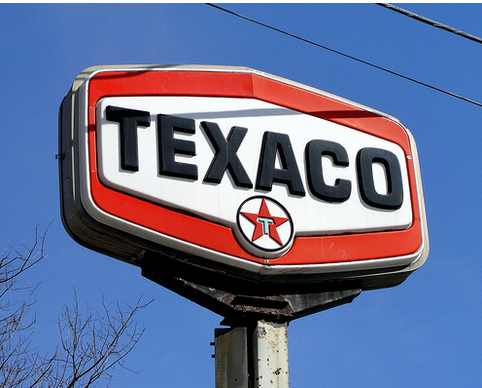}}}&
    \bmvaHangBox{\fbox{\includegraphics[height=20mm]{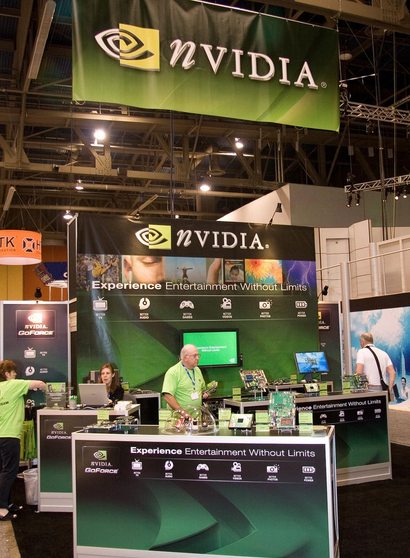}}} &
    \bmvaHangBox{\fbox{\includegraphics[height=20mm]{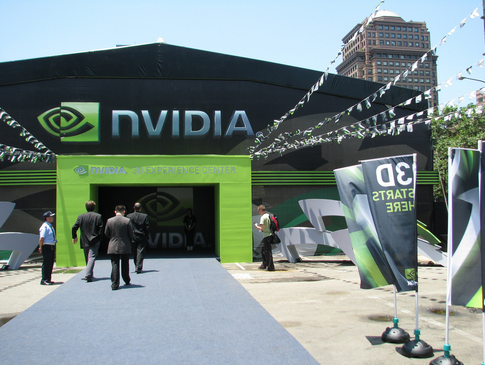}}}\\
    (a) & (b) & (c) & (d) \\
    Texaco Circle & Texaco Hexagon & $n$VIDIA old & NVIDIA new \\
  \end{tabular}
  \caption{{\bf Variations within categories.} Texaco training images have the circular logo (a), while some of the test images have the hexagonal Texaco logo (b). The script $n$VIDIA (c) is easier than the all-caps version (d). Correctly identified: (a), (c). Mistakes: (b), (d)}
  \label{fig:logo-versions}
\end{center}
\end{figure}

We defined several DCNN architectures in Section~\ref{sec:architectures}.
We apply these to logo classification, and we report accuracy results in Table~\ref{T:logo-classification}.
In the previous work, the most accurate classifier on FlickrLogos-32 is the Complete Rank Transform, which is built on top of SIFT features~\cite{Boia2014}.
GoogLeNet-GP produces higher logo classification accuracy than the Complete Rank Transform.
Increasing dropout boosts GoogLeNet logo classification accuracy.
Higher dropout also leads to slower convergence in training (Figure~\ref{fig:convergence}), but our highest-accuracy classifier trains in just fifteen minutes.

Although we are close to 90\% accuracy with GoogLeNet-GP, it is valuable to analyze the images that were incorrectly classified.
We see two key failure modes: low-resolution logos, and variation within logo categories.
First, we find that low-resolution logos are the most common failure case for all of our classifiers (GoogLeNet-GP, AlexNet, etc) -- we illustrate this pattern in Figures~\ref{fig:mistakes_bbox_size} and~\ref{fig:logo-scale}.
Second, logos change frequently.
Disney changed its logo more than 30 times from 1988 to 2015~\cite{DisneyLogo}.
In FlickrLogos-32, we found that substantially different versions of the Texaco logo appear in the trainval and test sets (see Figure~\ref{fig:logo-versions}).

Training protocol for {\bf GoogLeNet} experiments:
We begin by pretraining GoogLeNet on ImageNet.
Then, we re-initialize the softmax layers (cls1\_fc2, cls2\_fc2, and cls3\_fc) to randomized weights with 32 outputs (for FlickrLogos-32) instead of 1000 outputs (for ImageNet's 1000 categories).
Finally, we fine-tune GoogLeNet on FlickrLogos-32-val.

Training protocol for {\bf GoogLeNet-GP} experiments:
We begin with pretrained GoogLeNet weights.
Then, we modify GoogLeNet by inserting global pooling layers prior to fully-connected layers \{cls1\_fc1, cls2\_fc1, cls3\_fc\}, which appear after Inception3, Inception6, and Inception9, respectively.
In the off-the-shelf GoogLeNet, cls1\_fc1 and cls2\_fc1 take 4$x$4$x$channels input and cls3\_fc takes 1$x$1$x$channels input.
But, global pooling produces 1$x$1$x$channels input for these fully-connected layers.
So, we re-initialize cls1\_fc1, cls2\_fc1, and cls3\_fc with random weights and an input size of 1$x$1$x$channels.

Training protocol for {\bf Full-Inception} experiments:
We begin with pretrained GoogLeNet weights.
Then, for the first two conv layers, we randomly select filters to warp to smaller or larger sizes.
This way, we create inception layers out of pretrained model weights -- instead of retraining from scratch.

\FloatBarrier
\subsection{Logo Detection without Localization}
\label{sec:detection_without_localization}

The problem of logo detection without localization is defined here as "which logo, if any, is in this image?"

\begin{table*}[t!]
	\caption{FlickrLogos-32 {\bf non-localized detection} APs.}
	\label{T:logo-detection-nonlocalized}

	\centering
	\begin{tabular}{l|cccccccc|c}
		& adidas & aldi & apple & becks & bmw & carls & chim & coke & \\
		& corona & dhl & erdi & esso & fedex & ferra & ford & fost & \\
		& google & guin & hein & hp & milka & nvid & paul & pepsi & \\
		Method & ritt & shell & sing & starb & stel & texa & tsin & ups & mAP\\
		\hline
		& 44.0 & 67.6 & 80.6 & 72.1 & 79.2 & 56.5 & 73.0 & 55.9 & \\
		& 90.9 & 53.3 & 76.9 & 86.9 & 68.7 & 90.9 & 84.3 & 85.4 & \\
FRCN + & 81.2 & 86.4 & 73.3 & N/A & 46.0 & 52.5 & 97.6 & 37.5 & \\
AlexNet (ours) & 63.0 & 55.6 & 88.8 & 98.1 & 83.8 & 80.2 & 85.2 & 75.0 & 73.3 \\
	\end{tabular}
\end{table*}

\begin{figure}[htb]
	\centering
	\begin{tabular}{ccc}
		\bmvaHangBox{\fbox{\includegraphics[height=25mm]{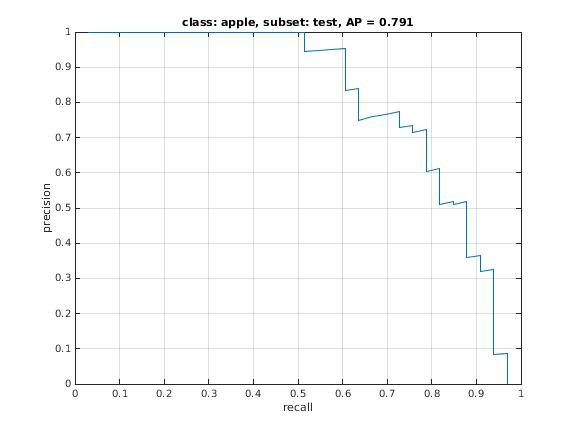}}}&
		\bmvaHangBox{\fbox{\includegraphics[height=25mm]{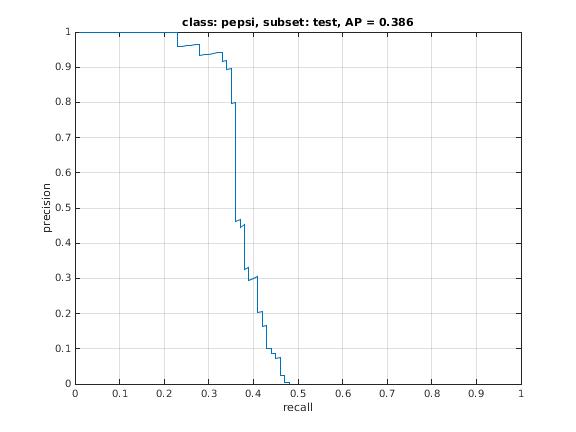}}}&
		\bmvaHangBox{\fbox{\includegraphics[height=25mm]{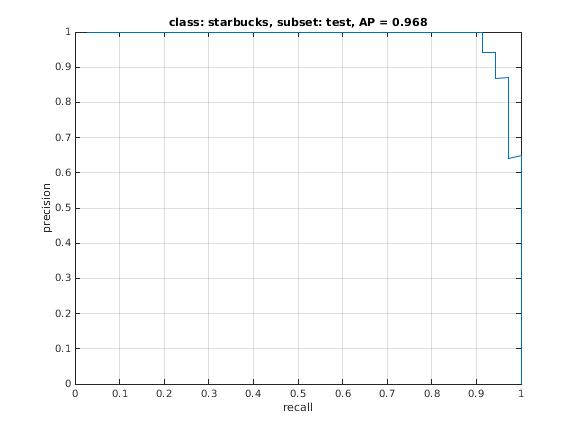}}} \\
		(a) & (b) & (c) \\
		Apple & Pepsi & Starbucks \\
	\end{tabular}
	\caption{{\bf Sample PR curves for Detection without Localization.} Detection without localization exhibits fairly good detection performance, especially on distinctive logos such as that of Starbucks.}
	\label{fig:flickrlogos_examples}
\end{figure}

We used Fast R-CNN (FRCN for short)~\cite{FastR-CNN} to do deep detection with localization of logos within images. FRCN is a framework for object detection built on top of Caffe ~\cite{jia2014caffe} that is significantly faster and more accurate than previous deep learning frameworks for object detection. It simplifies the training process for deep networks using a multi-task loss, achieving 9x faster training times than the previous version of R-CNN ~\cite{R-CNN}. These properties make Fast R-CNN a good choice of a framework to apply to our task.

Ordinarily, FRCN takes in {\em region proposals} and performs bounding box regression in order to refine these proposals. We will explore this functionality when we do localization with detection, but here, we used FRCN with only one region proposal per image (encompassing the entire image) and removed the bounding box regression functionality of FRCN. FRCN takes in this single region proposal and the image itself, which can be thought of as examining the entire image for logos. It then outputs a classification of the image as containing no logo, or, if a logo is present, which logo is in the image. We ran this modified FRCN with AlexNet and produced a 73.3\% mean average precision (more details in Table~\ref{T:logo-detection-nonlocalized}).
We are not aware of a baseline for non-localized detection in the related work -- FlickrLogos-32 (and other logo datasets) are most commonly used for evaluating retrieval algorithms.

\FloatBarrier
\subsection{Logo Detection with Localization}
\label{sec:detection_with_localization}
We then looked at the problem of detection with localization. Here, we define this problem as ``If there is a logo in this image, which logo is it and where is it located?"

Here, we use FRCN with {\em region proposals}. 
For both training and testing, FRCN takes in raw images and region proposals of significant features inside of those images in the form of bounding boxes. 
This is the localization step. 
It then classifies each bounding box region proposal as a type of logo or as ``background," meaning that the area in the proposal is not a logo at all. 
If the region proposal contains a logo, it also outputs a bounding box regression offset that adjusts the region proposal to more closely highlight the region containing the logo. 
Our region proposals are generated using selective search ~\cite{SelectiveSearch}. 
These regions along with images are then fed into FRCN for training/testing. 

As in the previous section, we do not know of a non-localized detection baseline.
We achieve a mean average precision of 73.5\% with FRCN+AlexNet and 74.4\% with FRCN +VGG16~\cite{VGG-19}. 
AP values per class and example precision-recall curves are shown in Table~\ref{T:logo-detection-localized} and Figure~\ref{fig:flickrlogos_examples}.

\begin{figure}[htb]
	\centering
	\begin{tabular}{ccc}
		\bmvaHangBox{\fbox{\includegraphics[height=25mm]{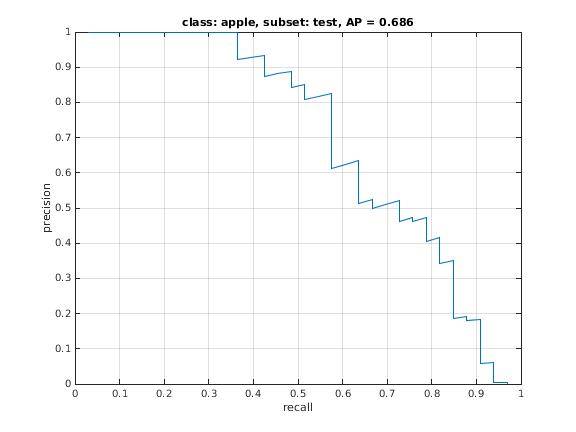}}}&
		\bmvaHangBox{\fbox{\includegraphics[height=25mm]{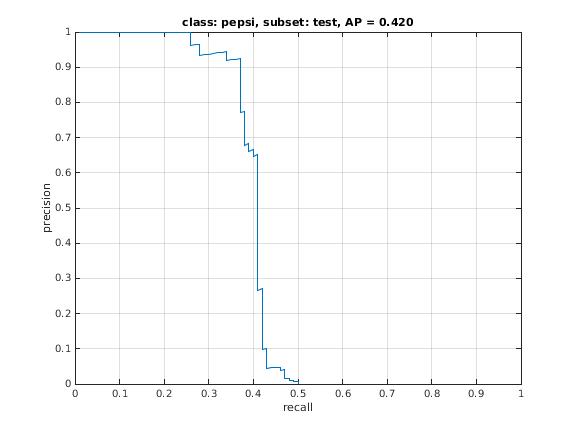}}}&
		\bmvaHangBox{\fbox{\includegraphics[height=25mm]{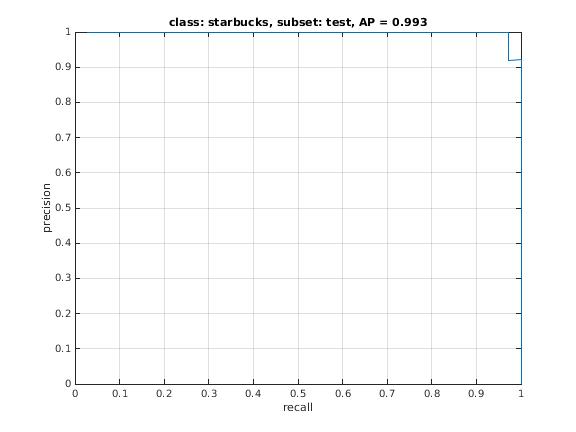}}} \\
		(a) & (b) & (c) \\
		Apple & Pepsi & Starbucks \\
	\end{tabular}
	\caption{{\bf Sample PR curves for Detection with Localization.} Detection with localization exhibits even better detection performance than detection without localization.}
	\label{fig:flickrlogos_examples}
\end{figure}

\begin{table*}[t!]

\caption{FlickrLogos-32 {\bf localized detection} APs.}
\label{T:logo-detection-localized}
\centering
\begin{tabular}{l|cccccccc|c}
 & adidas & aldi & apple & becks & bmw & carls & chim & coke & \\
 & corona & dhl & erdi & esso & fedex & ferra & ford & fost & \\
 & google & guin & hein & hp & milka & nvid & paul & pepsi & \\
Method & ritt & shell & sing & starb & stel & texa & tsin & ups & mAP\\
\hline
 & 47.8 & 69.1 & 68.6 & 71.7 & 81.7 & 59.1 & 70.9 & 58.8 & \\
 & 90.9 & 56.9 & 83.6 & 89.4 & 70.8 & 88.3 & 85.7 & 86.0 & \\
FRCN + & 90.9 & 81.0 & 66.5 & N/A  & 46.6 & 52.4 & 98.0 & 42.0 & \\
AlexNet (ours) & 63.3 & 50.6 & 83.5 & 99.3 & 87.9 & 81.5 & 86.7 & 70.5 & 73.5 \\
\hline
 & 61.6 & 67.2 & 84.9 & 72.5 & 70.0 & 49.6 & 71.9 & 33.0 & \\
 & 92.9 & 53.5 & 80.1 & 88.8 & 61.3 & 90.0 & 84.2 & 79.7 & \\
FRCN + & 85.2 & 89.4 & 57.8 & N/A  & 34.6 & 50.3 & 98.6 & 34.2 & \\
VGG16 (ours) & 63.0 & 57.4 & 94.2 & 95.9 & 82.2 & 87.4 & 84.3 & 81.5 & 74.4
\\
\end{tabular}
\end{table*}
\vspace{-0.2in}

\section{Conclusion}
\label{sec:conclusion}
Logo recognition is a key problem in marketing analytics, digital advertising, and augmented reality.
We have developed deep convolutional neural network (DCNN) architectures that improve upon state-of-the-art logo classification results.
We also use our DCNNs to establish a baseline for accuracy on logo detection.
Collectively, we refer to this work as {\em DeepLogo}.

\section*{Acknowledgements}
Research partially funded by DARPA Award Number HR0011-12-2-0016, plus ASPIRE industrial sponsors and affiliates Intel, Google, Huawei, Nokia, NVIDIA, Oracle, and Samsung. 
The first author is funded by the US DOD NDSEG Fellowship.

\bibliography{bibliography}
\end{document}